\documentclass{article}

\usepackage{arxiv}

\usepackage[utf8]{inputenc} 
\usepackage[T1]{fontenc}    
\usepackage{hyperref}       
\usepackage{url}            
\usepackage{booktabs}       
\usepackage{amsfonts}       
\usepackage{nicefrac}       
\usepackage{microtype}      
\usepackage{mwe}
\usepackage{pifont}
\usepackage{mathdots}
\usepackage{amsmath}
\usepackage{caption} 
\title{Combining word embeddings and convolutional neural networks to detect duplicated questions}

\author{
    Yoan Dimitrov \\
    Department of Computer Science \\
    Reutlingen University \\
    Pestalozzistraße 62, 72762 Reutlingen, Germany \\
  \texttt{yoan.dimitrov@student.reutlingen-university.de}}
  
\begin{document}
\maketitle

\begin{abstract}
Detecting semantic similarities between sentences is still a challenge today due to the ambiguity of natural languages. In this work, we propose a simple approach to identifying semantically similar questions by combining the strengths of word embeddings and Convolutional Neural Networks (CNNs). In addition, we demonstrate how the cosine similarity metric can be used to effectively compare feature vectors. Our network is trained on the Quora dataset, which contains over 400k question pairs. We experiment with different embedding approaches such as Word2Vec, Fasttext, and Doc2Vec and investigate the effects these approaches have on model performance. Our model achieves competitive results on the Quora dataset and complements the well-established evidence that CNNs can be utilized for paraphrase detection tasks.
\end{abstract}


\section{Introduction}
Over the past decade, numerous online community forums such as Stack Overflow and Quora have emerged, enabling users to ask domain-specific questions and obtain appropriate answers. Since many users share the same interests and often ask the same questions, these online forums face the challenge of identifying semantically similar or equivalent questions in order to avoid redundancies and improve their service quality and user experience \cite{Bo15}. Successfully measuring the semantic similarity between pairs of questions is, however, a very difficult hurdle to overcome, since natural languages are ambiguous at all major levels of linguistics. Two questions could vary in length, punctuation, sentence structure, and choice of words and still have the same intention. For example, consider the following two questions:  \textit{How old are you?}  \textit{What is your age?} Both questions have the same intention and can be answered with the same answer, but they both have a different vocabulary and sentence structure. These aspects, although seemingly trivial, are not always easy to detect. Recent years have seen great progress in detecting semantically similar questions with deep learning. In particular, Recurrent Neural Networks (RNNs) \cite{HSY17, NM17, Sh18, WHF17}, which are inherently ideal for sequential data such as text, as well as attention mechanisms \cite{GLZ18, To17, Yang19} have achieved state-of-the-art results. 

In this work, we tackle the task of identifying semantically similar questions from the officially released Quora dataset\footnote{Quora Dataset: https://www.quora.com/q/quoradata/First-Quora-Dataset-Release-Question-Pairs}. We adopt the question similarity definition suggested by Bogdonova et al. \cite{Bo15}: 
"\textit{Two questions are considered to be semantically equivalent if they can be answered by the same answer.}"

Unlike more popular approaches, we propose a simple but effective method based solely on word embeddings and a Convolutional Neural Network (CNN) architecture introduced by Yoon Kim \cite{Ki14}. Our experimental results are comparable to more complex state-of-the-art models and also shed light on the strengths of word embeddings as well as the importance of cleaning noisy text data.

The outline of the paper is as follows. Section 2 gives a brief overview of related work on the detection of semantically equivalent questions. All details of our proposed model as well as the necessary data pre-processing steps are presented in sections 3 and 4. Our model is evaluated in Section 5 and compared with other approaches. Furthermore, the limitations of our model and future research directions are discussed. Lastly, concluding remarks are made in section 6.

\section{Related Work}
There have been numerous academic contributions dealing with paraphrase detection and the detection of semantically equivalent questions. The following section gives an overview of recent research conducted with the Quora dataset. 

Wang et al. \cite{WHF17} were the first to publish impressive results on the detection of duplicated questions with the Quora dataset. They proposed a bilateral multi-perspective matching (BiMPM) model, which compares sentence embeddings P and Q from multiple perspectives and in both directions ($P \rightarrow Q$ and $P \leftarrow Q$). They create contextual embeddings for each question with a bi-directional Long Short-Term Memory Network (LSTM). The authors achieved an impressive 88.17\% classification accuracy. 

Tomar et al. \cite{To17} took on a self-attention approach and used character-level n-gram embeddings and word embeddings to encode question pairs. They pretrained their decomposable attention (DECATT) model with noisy data from other domains in order to improve overall performance. They managed to obtain an accuracy of 88.40\%. 

Homma et al. \cite{HSY17} employed a Siamese Gated Recurrent Unit (GRU) and applied data augmentation to their model for better results. They achieved 85\% accuracy on the Quora dataset. Furthermore, they also experimented with the following distances metrics to calculate the similarity between sentence vectors: Euclidian distance, Cosine distance, and weighted Manhatten distance. According to their studies, the weighted Manhatten distance performed best. 

A relatively unique approach for detecting duplicated questions from the Quora dataset was demonstrated in \cite{De18} by Deudon. The author uses a variational autoencoder to derive the intentions of the questions as normal distributions. He argues that decomposing the representation of a sentence in a mean vector and a diagonal covariance matrix accounts for uncertainty and ambiguity in natural languages and is therefore more appropriate than single vectors which capture relationships between multiple words and phrases. The Gaussian intents are fed into a Variational Siamese network and compared using a Wasserstein distance metric. He recorded an accuracy of 88.86\% on the Quora dataset. 

Gong et al. \cite{GLZ18} introduced a Densely Interactive Inference Network (DIIN) to achieve high-level text comprehension by utilizing syntactic features and hierarchically extracting semantic features from interaction space. They make use of a CNN to acquire important features, which are then aggregated and passed to an output layer. The output layer predicts a confidence level for each class. Their model achieved state-of-the-art performance on multiple, well known datasets. On the Quora dataset, they were able to achieve an 89\% accuracy.

Nicosia et al. \cite{NM17} experimented with RNNs and CNNs as sentence encoders and compared two sentence representations by minimizing a contrastive loss function based on the euclidean distance. Their best model on the Quora dataset was a GRU, which achieved an 86,82\% accuracy.

Rao et al. \cite{Ra19} proposed a hybrid CNN/LSTM co-attention model (HCAN), which combines semantic matching with relevance
matching. They achieved 85.3\% accuracy on the Quora dataset.

Shen et al. \cite{Sh18} conducted a comparative study between Simple Word-Embedding based Models (SWEMs) and RNN/CNN based models.  Their aim was to understand when and why simple pooling strategies such as max-pooling and hierarchical pooling can be applied to word embeddings without any further processing. They found that such strategies are sufficient in some cases of natural language understanding. They experimented with 17 different datasets. Their best model on the Quora dataset achieved an accuracy of 83.03\%.

Yang et al. \cite{Yang19} proposed an efficient, text-matching model named RE2, which uses previously aligned features (residual
vectors), pointwise features (embedding vectors) and contextual features (encoded vectors) to obtain useful information from texts. Their model, which relies on convolutional layers to extract contextual features and attention to compare features, achieved on average an accuracy of 89.2\% on the Quora dataset.

\section{Approach}
In the following sections, we present the implementation details and background of our proposed model. We particularly focus on the data pre-processing pipeline and the relevant operations and layers of our network architecture.

\subsection{Problem Formulation}

Let $P = (p_1,…,p_{M} )$ and $Q = (q_1,…,q_{N} )$ be two questions consisting of $M$ and $N$ words, respectively. Furthermore, let $y \in \{0,1\}$ be the binary label indicating whether questions P and Q are duplicated or not. Our model’s goal is to predict the y label that best reflects the semantic relationship between P and Q. As Wang et al. point out \cite{WHF17}, the problem can also be seen as the estimation of the conditional probability $P(y | P,Q)$.

\subsection{Text Pre-Processing}

The Quora dataset consists of 404,351 question pairs with positive and negative labels indicating whether or not two questions are duplicated. Roughly 37\% of the question pairs are duplicates. The question pairs go through several pre-processing steps before being converted into numerical representations. First, every question is lower-cased and stripped of all punctuation and other non-alphanumeric characters. Then each question is tokenized with the Natural Language Toolkit (NLTK) tokenizer \cite{BKL09}.  We decide not to remove stop words as they are considered to be semantically important in many cases. Furthermore, we do not discard rare words that occur only once or twice since they make up 43.21\% of the text corpus. To train our neural network, all input questions need to have a fixed length. Therefore, we clip all sentences longer than 40 words. We decided on the length 40 after finding out that less than 1\% of the questions contain more than 40 tokens. 

In an attempt to fix numerous spelling errors which could negatively affect the performance of the model, we employ a pre-trained Word2Vec model \cite{Mi13a, Mi13b} and Symspellpy\footnote{SymSpell: https://github.com/wolfgarbe/SymSpell}. The Word2Vec model was trained on a Wikipedia dataset and therefore has a huge vocabulary. It is assumed that words which cannot be found by the Word2Vec model either contain spelling errors or are unknown terms. Such words are passed to Symspellpy, a tool that corrects spelling errors and segments words accordingly if necessary. Words, which Symspellpy cannot correct, are not discarded. We assume that such words are unknown names of people or simply outside the vocabulary (OOV). A similar approach can be found in \cite{HSY17}.

\subsection{Word Embeddings}

The preprocessed questions are transformed into numerical representations before being used as input to our neural network. We experiment with three different embedding approaches: Word2Vec \cite{Mi13a, Mi13b}, Fasttext \cite{Bo14, Jo17} and Doc2Vec \cite{LM14}. In addition, we introduce a fourth approach by combining Word2Vec and Term Frequency-Inverse Document Frequency (TF-IDF) \cite{Ra03} \cite [pp.~105-106]{JM18}. 

Word2Vec was introduced by Mikolov et al. \cite{Mi13a, Mi13b} in 2013. The authors proposed two closely related neural network architectures for learning word vector representations: Continous Bag-Of-Words (CBOW) and Skip-Gram. The CBOW model attempts to learn the vector representation of a target word by taking some surrounding context words into account, while the Skip-Gram model learns the vector representations of some corresponding context words for a given target word.

Bojanowski et al. \cite{Bo14} introduced Fasttext, an extension to Word2Vec, after recognizing that rare words are often poorly estimated or even regarded as insignificant to some algorithms. Instead of only learning vector representations on whole words in a text corpus like Word2Vec does, Fasttext is capable of learning the n-grams within a word \cite{Bo14, Jo17}. This, in turn, allows Fasttext to be more sensitive to OOV and rare words. The model first splits words into n-grams and then feeds the n-grams to a CBOW or Skip-Gram model in order to learn the embeddings.

Doc2Vec, also considered to be an extension to Word2Vec, was introduced by Le and Mikolov in 2014 \cite{LM14}. This model attempts to learn how to map both words and entire sentences to an embedding space while capturing important semantics. 

In this work, Word2Vec, Fasttext, and Doc2Vec are trained using the Skip-Gram architecture. This is due to the fact that Skip-Gram works well on smaller datasets and is more susceptible to rare words \cite{Mi13a, Mi13b}. Each model is trained for 30 epochs with the same parameters and generates 300-dimensional word vectors.

For the fourth embedding approach, we first calculate the TF-IDF score of every word in the text corpus. Frequent words like “the, a, and, or”, which are found across all sentences, are penalized; simultaneously rare words across the whole text corpus receive high scores \cite{Ra03} \cite [pp.~105-106]{JM18}. We then iterate through the vocabulary $V$, which represents all unique words in the dataset, and multiply the vector representation $w$ of each word with its corresponding $tf\_idf$-score as shown in Equation 1. The result is an embedding matrix  $E \in \mathbb{R}^{n \times d}$, the rows of which are the TF-IDF-weighted word vectors across 300 dimensions.

\begin{equation}
  E = \sum_{i=0}^V w_i \times tf\_idf_i
\end{equation}

\subsection{Network Architecture}

The network used to classify question pairs as duplicates or non-duplicates is largely based on the CNN architecture for sentence classification introduced by Yoon Kim \cite{Ki14} in 2014. 

We begin with an embedding layer, which maps all words in a pair of questions $\{P,Q\}$ to their corresponding d-dimensional vector representations. This results in the matrices 
$P \in \mathbb{R}^{n\times d}$ and $Q \in \mathbb{R}^{n\times d}$, where $n$ is the maximum (null-padded) sentence length, and $d$ is the dimensionality of the vector representations \cite{Ki14, ZW15, JSG18}. Note that the word vector representations could come from any of the four embedding approaches mentioned previously.  The matrices $P$ and $Q$ are then fed to separate 1d convolutional layers to extract different levels of semantic features for each question. To avoid repetition we will only explain how features are extracted from the matrix $P$, since the exact same procedure is used for matrix Q.

Given the matrix $P$, whose rows consist of a sequence of word representations, we denote $P[i{:}j]$ as a sub-matrix of $P$ from row $i$ to row $j$ \cite{Ki14}. In order to extract features, we apply a filter $w \in \mathbb{R}^{h\times k}$ to a window of $h$ words in the matrix $P$, while the width $k$ of the filter remains fixed \cite{Ki14, ZW15, JSG18}. 
A feature map $c_i\in \mathbb{R}^{n-h+1}$ is generated by repeatedly applying the filter $w$ on the sub-matrices of $P$:

\begin{equation}
  c_i=\alpha(w\cdot P[i{:}i+h-1]+b)  
\end{equation}

Here, the $\cdot$ symbol represents the dot product, $b \in \mathbb{R}$ is a bias term, and $\alpha()$ is a non-linear activation function \cite{Ki14, ZW15}. In this work, we use the element-wise Rectified Linear Unit (ReLU) activation function. Generally, we apply five filter windows \{2, 3, 4, 6, 8\} 200 times to matrices $P$ and $Q$. This means each convolution layer outputs a matrix $c \in \mathbb{R}^{(n-h+1) \times 200}$. After generating the feature maps for all corresponding filter sizes, we then apply a global max-pooling operation (Eq. 3) as well as a global min-pooling operation (Eq. 4) over each matrix $c$ \cite{Ki14}. Furthermore, we split the matrix $c$ into $i$ equal chunks, apply the global max-pooling operation to each chunk, and then concatenate their max-values (Eq. 5). A similar method was used by Liu et al. \cite{Li17} for text classification.

\begin{equation} 
\begin{aligned}
{c_{max}= max(c)} \\
with \: c_{max}\in \mathbb{R}^{1\times 200}
\end{aligned}
\end{equation}

\begin{equation} 
\begin{aligned}
{c_{min}= min(c)} \\
with\: c_{min}\in \mathbb{R}^{1\times 200}
\end{aligned}
\end{equation} 

\begin{equation} 
{c_{max_i}= max({z_1})\oplus\ldots\oplus max({z_i})}
\end{equation}

where $z_1,\dots z_i$ represent four equal sub-matrices $\in \mathbb{R}^{(n-h+1) \times 50}$ of the matrix $c$. The symbol $\oplus$ is used to denote the concatenation operator. As Kim mentions in \cite{Ki14}, max-pooling captures the most important feature for every feature map. In the same manner, the min-pooling operation captures the least important feature for every feature map. We include this feature as well as the $c_{max_i}$ features, as they capture additional important information about each question.

In order to find out if matrices $Q$ and $P$ are semantically equivalent or not, we compare their respective $c_{max}$, $c_{min}$, and $c_{max_i}$ features, which are extracted from each feature map. More formally, we calculate their cosine similarities:

\begin{equation} 
{cos(\theta) = \frac{f_1 \cdot f_2}{\|f_1\| \times \|f_2\|}} + \lambda
\end{equation}

where $f_1$ and $f_2$ represent the corresponding $c_{max}$, $c_{min}$ or $c_{max_i}$  values for matrices $Q$ and $P$ and \begin{math} \lambda \end{math} is an additional, trainable parameter with a range from 0 to 1. However, its initial value is 1. Recall that our model uses five filters with varying window sizes to obtain multiple features for each question. Therefore, we compare the features $f_1$ and $f_2$  filter-wise in both directions\footnote{ The $ \lambda$ term was introduced as a trainable parameter, which can influence the similarity score, because the dot product is commutative ($\vec{a}\cdot\vec{b} = \vec{b}\cdot\vec{a}$)} ($f_1$$\rightarrow$$f_2$ and $f_2$$\leftarrow$$f_1$) \cite{WHF17}. After matching all feature vectors for Q and P, we concatenate their cosine similarity values as follows:

\begin{equation} 
{x= x_1\oplus x_2 \oplus\ldots\oplus x_n}
\end{equation}
The concatenated vector $x$ is then passed through a softmax function, which produces the distribution over the classification classes $\{0,1\}$ from which the strongest class is outputted.

\subsection{Network Inputs}
Originally, we designed the CNN architecture to compare two questions in the form of $d \times n$ matrices. After some experimentation, however, we decided to also explicitly include the unique words $\{P_u,Q_u\}$ and keywords $\{P_k,Q_k\}$ of each question pair as additional, embedded inputs to the network, resulting in three inputs for each question. 

\begin{figure}[!ht]
  \centering
  \includegraphics[width=.64\textwidth]{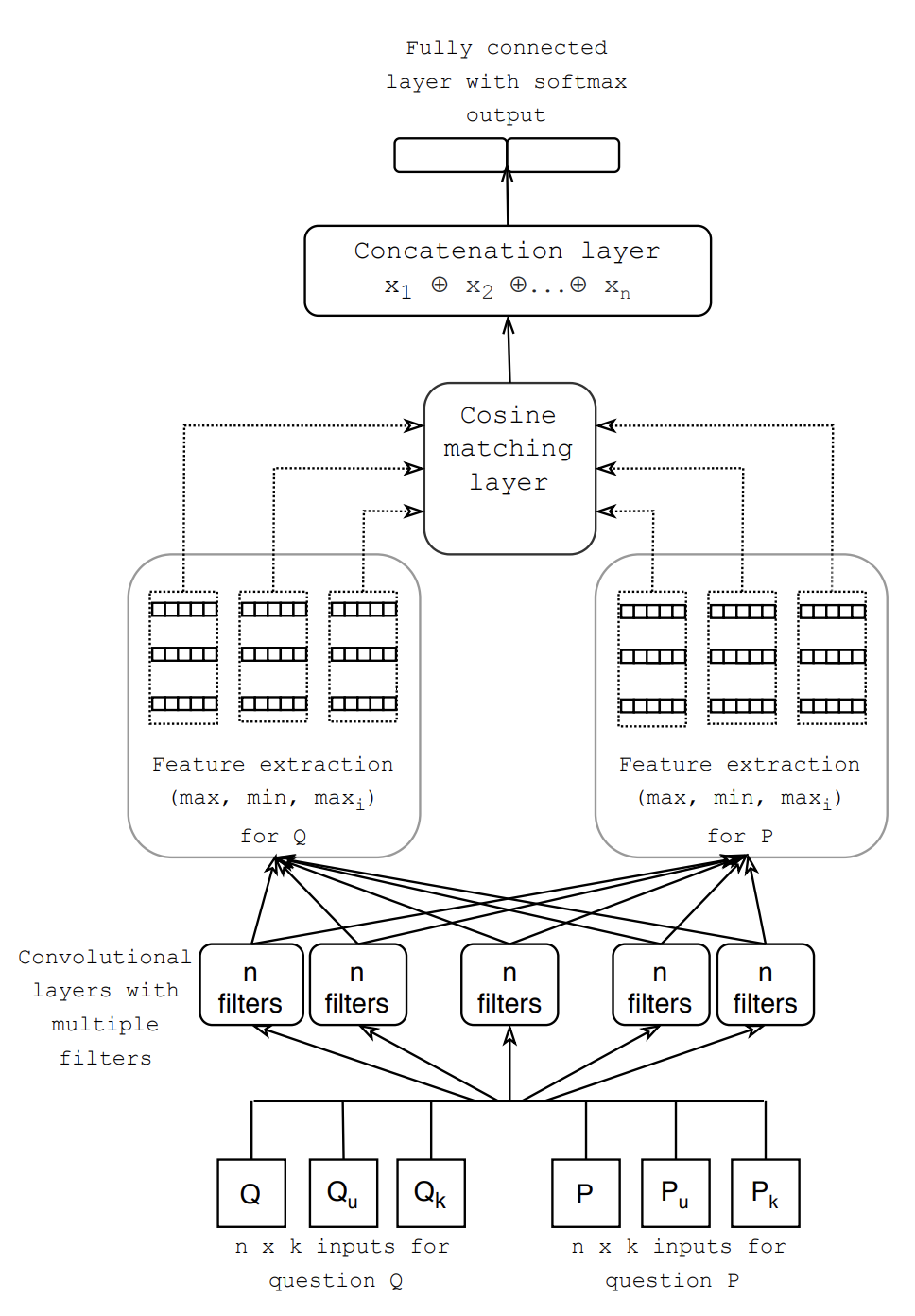}
  \caption{Proposed CNN architecture}
  \label{Figure:1}
\end{figure}

We apply convolutional filters to the pairs $\{P,Q\}$, $\{P_u,Q_u\}$, and $\{P_k,Q_k\}$ and extract their $max$, $max_i$, and $min$ features. Thereafter, we compare the features of $P$,  $P_k$, and $P_u$ with the features of $Q$,  $Q_k$, and $Q_u$ filter-wise using the cosine similarity function (Eq. 6). The resulting vectors are then concatenated and fed to the softmax function to generate the final prediction. This approach is depicted in Figure 1. It should be noted that activation and regularization layers are omitted to avoid further complexity.

In order to find the unique words for every question, we simply retrieve the words that do not intersect in a question pair. The keyword extraction from each question is based on the words’ $tf\_idf$-scores. The three words with the highest $tf\_idf$-scores in a question are selected as keywords. Since frequent words are penalized by the TF-IDF-algorithm and therefore have a lower score, they are usually not included as key words.

\section{Parameters and Training}
For all experiments, we train with a mini-batch size of 64 and use the non-linear ReLU activation function on all convolutional layers. All the weights of these layers are randomly initialized with the Xavier initializer \cite{GB10}. To avoid overfitting and stabilize training, we apply a dropout  rate \cite{Sr14} of 0.1 after every convolutional layer and use batch normalization \cite{IS15} with a momentum of 0.7 after every embedding layer.

Our model is trained and evaluated on the exact same data partitions that Wang et al. published \cite{WHF17}. This allows us to compare our results fairly with theirs and other published results \cite{De18,GLZ18,HSY17, NM17, Sh18, To17, Yang19}, which also use the same data partitions. The dataset is split into three partitions: 10,000 question pairs for validation, 10,000 question pairs for testing, and the rest for training.

Our model’s goal is to minimize the binary cross entropy of the predicted and true distributions. We employ the ADAM optimizer \cite{KB15} with $lr = 0.001,\beta1 = 0.9,\beta2 = 0.999$ to update model parameters. The learning rate $lr$ is decayed after every second epoch by a factor of 0.1 if the validation loss has not decreased. During training, we also update the pre-trained word embeddings for the pairs $\{P,Q\}$ and $\{P_u,Q_u\}$. The weights of the keywords pair $\{P_k,Q_k\}$, are left static during training \cite{Ki14}, since they do not improve model performance. 

As we train our model with each of the four embedding approaches, we select the models that perform well on the validation set and then evaluate them on the test set. Accuracy is used as the evaluation metric for the dataset. The training is aborted if there is no update to the best accuracy on the validation set for the last 3 epochs.

\section{Experimental Results}
In Table \ref{tab:results} we compare our baselines with other models trained on the Quora dataset. Each model’s validation accuracy (if published) and test accuracy are listed in the columns. The models are sorted by decreasing test accuracy. Note that all models with \ding{169} were reported by Wang et al. in their work \cite{WHF17}. Our baselines are denoted in bold in the last four rows of the table. Each baseline is labeled by the embedding approach used during training. Our experimental results surpass some models in terms of test accuracy and are also competitive with the state-of-the-art models. Among our models, we found Word2Vec to be the most consistent embedding approach. 

\begin{table}
\centering
\begin{tabular}{lll}
\toprule
Model & Val. Accuracy & Test Accuracy \\
\midrule
RE2 \cite{Yang19}  & - & 89.2 \\
DIIN \cite{GLZ18}	& 89.44	 & 89.06\\
VAR-Siamese \cite{De18}	& 89.05	& 88.86\\
DECATT \cite{To17}	& 88.89	& 88.40\\
BiMPM \cite{WHF17}	& 88.69	& 88.17\\
Contrastive-GRU \cite{NM17} & - & 86.82 \\
LCD \ding{169} & - & 85.55 \\
HCAN  \cite{Ra19} & - & 85.3 \\
Siamese-GRU \cite{HSY17} & - & 85.0 \\
Multi-Perspective-LSTM \ding{169} & - & 83.21 \\
SWEM-concat \cite{Sh18} & - & 83.03 \\
Siamese-LSTM \ding{169} & - & 82.58 \\
Multi-Perspective-CNN \ding{169} & - & 81.38 \\
Siamese-CNN \ding{169} & - & 79.60 \\
\textbf{W2V} & \textbf{87.66} & \textbf{87.22} \\
\textbf{FAST} & \textbf{87.10} & \textbf{87.02} \\
\textbf{D2V} & \textbf{87.35} & \textbf{86.92} \\
\textbf{W2V+TFIDF} & \textbf{87.03} & \textbf{86.58} 
\end{tabular}
\caption{Performance comparison of different models on the Quora dataset}
\label{tab:results}
\end{table}

\begin{table}
\centering
\begin{tabular}{|p{4cm}|p{4cm}|p{0.5cm}|p{0.5cm}|}

\toprule
Q1 & Q2 & P & T \\
\midrule
How do I get funding for my web-based startup idea? &	How do I get seed funding pre-product? &	1 &	1 \\ 
When did the first world war happen? &	What were the causes of World War I? &	0 &	0 \\ 
How deep is the Titanic wreck? & How deep is Atlantic Ocean where Titanic sank? &	0 &	1 \\ 
What is the physical meaning of divergence, curl and gradient of a vector field? &	Why curl represents rotation of a vector field? &	0 &	1 \\ 
What is definition of instrumentation? & What is biomedical instrumentation? &	1	& 0
\end{tabular}
\caption{Model prediction examples}
\label{tab:predictions}
\end{table}

In Table \ref{tab:predictions}  we show several correct and incorrect label predictions on the validation set from our best model. The P column represents the predicted binary label, while the T column represents the true binary label from the validation set. The first two rows contain pairs of questions that our model classified correctly. The last three rows show questions that were incorrectly classified by our model.

During the training of our models we observed two interesting, behavioral patterns: A model sometimes tends to correctly predict most non-duplicated pairs of questions, but at the same time has difficulties in predicting duplicates. The reverse case can also occur, which makes it even more challenging to find an optimal equilibrium. In addition, we found that our models often predict question pairs correctly from a human point of view, but incorrectly according to the dataset. Further investigation revealed that the Quora dataset has many question pairs that are labeled with a high degree of subjectivity. This evidently affects the performance of our model and its ability to generalize, since it depends primarily on word embeddings.

With regard to future improvements of our approach, we believe using character-level embeddings \cite{ZZL15} instead of word embeddings or even combing both methods could boost performance. Combining domain-specific word embeddings with generic embeddings has also been shown to improve performance \cite{LC16}. Future work may also involve hyper-parameter tuning as well as the study of model ensemble techniques, other approaches for extracting features from convolutional layers, and other similarity measures for comparing features. 
\section{Conclusion}
We presented a simple, effective CNN approach built on top of word embeddings to detect duplicated questions from the Quora dataset. Correcting spelling errors in the data set with the pre-trained Word2Vec model improved our model performance by 0.2\% on average. We employed four different embedding approaches, which all led to similar results. Using global max-pooling, min-pooling and max-pooling on chunks to extract important features from each convolutional layer proved to be very effective. We also experimented with the average-pooling operation, but it performed poorly for our task. Overall, our competitive results show the potential of CNNs for the detection of semantically equivalent questions, a task that is typically solved by using RNNs and / or attention mechanisms.

\bibliographystyle{acm}  
\bibliography{references}

\begin{thebibliography}{10}

\bibitem{BKL09}
{\sc Bird, S., Klein, E., and Loper, E.}
\newblock {\em Natural language processing with Python: analyzing text with the
  natural language toolkit}.
\newblock O’Reilly Media, Inc., 2009.

\bibitem{Bo15}
{\sc Bogdanova, D., dos Santos, C., Barbosa, L., and Zadrozny, B.}
\newblock Detecting semantically equivalent questions in online user forums.
\newblock In {\em Proceedings of the Nineteenth Conference on Computational
  Natural Language Learning\/} (Beijing, China, 2015), Association for
  Computational Linguistics, pp.~123--131.

\bibitem{Bo14}
{\sc Bojanowski, P., Grave, E., Joulin, A., and Mikolov, T.}
\newblock Enriching word vectors with subword information.
\newblock In {\em Transactions of the Association for Computational
  Linguistics\/} (2014), vol.~5, p.~1929–1958.

\bibitem{De18}
{\sc Deudon, M.}
\newblock Learning semantic similarity in a continuous space.
\newblock In {\em Advances in Neural Information Processing Systems 31\/}
  (2018), Curran Associates, Inc., pp.~986--997.

\bibitem{GB10}
{\sc Glorot, X., and Bengio, Y.}
\newblock Understanding the difficulty of training deep feedforward neural
  networks.
\newblock In {\em Proceedings of the thirteenth international conference on
  artificial intelligence and statistics\/} (2010), pp.~249--256.

\bibitem{GLZ18}
{\sc Gong, Y., Luo, H., and Zhang, J.}
\newblock Natural language inference over interaction space.
\newblock In {\em International Conference on Learning Representations\/}
  (Vancouver, Canada, 2018).

\bibitem{HSY17}
{\sc Homma, Y., Sy, S., and Yeh, C.}
\newblock Detecting duplicate questions with deep learning.
\newblock In {\em Stanford CS224n report\/} (2017).

\bibitem{IS15}
{\sc Ioffe, S., and Szegedy, C.}
\newblock Batch normalization: Accelerating deep network training by reducing
  internal covariate shift.
\newblock In {\em Proceedings of the 32nd International Conference on Machine
  Learning\/} (Lille, France, 2015), F.~Bach and D.~Blei, Eds., vol.~37 of {\em
  Proceedings of Machine Learning Research}, {PMLR}, pp.~448--456.

\bibitem{JSG18}
{\sc Jacovi, A., Shalom, O., and Goldberg, Y.}
\newblock Understanding convolutional neural networks for text classification.
\newblock In {\em Proceedings of the 2018 EMNLP Workshop Blackbox NLP:
  Analyzing and Interpreting Neural Networks for NLP\/} (Brussels, Belgium,
  2018), Association for Computational Linguistics, pp.~56--65.

\bibitem{Jo17}
{\sc Joulin, A., Grave, E., Bojanowski, P., and Mikolov, T.}
\newblock Bag of tricks for efficient text classification.
\newblock In {\em Proceedings of the 15th Conference of the {E}uropean Chapter
  of the Association for Computational Linguistics: Volume 2, Short Papers\/}
  (Valencia, Spain, 2017), Association for Computational Linguistics,
  pp.~427--431.

\bibitem{JM18}
{\sc Jurafsky, D., and Martin, J.~H.}
\newblock {\em Speech and Language Processing: An Introduction to Natural
  Language Processing, Computational Linguistics, and Speech Recognition.}
\newblock Pearson. Prentice Hall, Third Edition draft, 2018.

\bibitem{Ki14}
{\sc Kim, Y.}
\newblock Natural language inference over interaction space.
\newblock In {\em Proceedings of the 2014 Conference on Empirical Methods in
  Natural Language Processing ({EMNLP})\/} (Doha, Qatar, 2014), Association for
  Computational Linguistics, pp.~1746--1751.

\bibitem{KB15}
{\sc Kingma, D.~P., and Ba, J.}
\newblock Adam: A method for stochastic optimization.
\newblock In {\em International Conference on Learning Representations
  {(ICLR)}\/} (San Diego, USA, 2015).

\bibitem{LM14}
{\sc Le, V.~Q., and Mikolov, T.}
\newblock Distributed representations of sentences and documents.
\newblock In {\em Proceedings of the 31st International Conference on
  International Conference on Machine Learning - Volume 32\/} (Beijing, China,
  2014), JMLR.org, pp.~1188--1196.

\bibitem{LC16}
{\sc Limsopatham, N., and Collier, N.}
\newblock Modelling the combination of generic and target domain embeddings in
  a convolutional neural network for sentence classification.
\newblock In {\em Proceedings of the 15th Workshop on Biomedical Natural
  Language Processing\/} (Berlin, Germany, 2016), Association for Computational
  Linguistics, pp.~136--140.

\bibitem{Li17}
{\sc Liu, J., Chang, W.-C., Wu, Y., and Yang, Y.}
\newblock Deep learning for extreme multi-label text classification.
\newblock In {\em Proceedings of the 40th International ACM SIGIR Conference on
  Research and Development in Information Retrieval\/} (New York, NY, USA,
  2017), Association for Computing Machinery, pp.~115--124.

\bibitem{Mi13a}
{\sc Mikolov, T., Chen, K., Corrado, G.~S., and Dean, J.}
\newblock Efficient estimation of word representations in vector space.
\newblock In {\em ICLR Workshop Papers\/} (2013).

\bibitem{Mi13b}
{\sc Mikolov, T., Sutskever, I., Chen, K., Corrado, G., and Dean, J.}
\newblock Distributed representations of words and phrases and their
  compositionality.
\newblock In {\em Advances in Neural Information Processing Systems 26\/}
  (2013), C.~J.~C. Burges, L.~Bottou, M.~Welling, Z.~Ghahramani, and K.~Q.
  Weinberger, Eds., Curran Associates, Inc., pp.~3111--3119.

\bibitem{NM17}
{\sc Nicosia, M., and Moschitti, A.}
\newblock Accurate sentence matching with hybrid siamese networks.
\newblock In {\em Proceedings of the 2017 ACM on Conference on Information and
  Knowledge Management\/} (New York, NY, USA, 2017), Association for Computing
  Machinery, pp.~2235--2238.

\bibitem{Ra03}
{\sc Ramos, J.}
\newblock Using tf-idf to determine word relevance in document queries.
\newblock In {\em Computer Science\/} (2003).

\bibitem{Ra19}
{\sc Rao, J., Liu, L., Tay, Y., Yang, W., Shi, P., and Lin, J.}
\newblock Bridging the gap between relevance matching and semantic matching for
  short text similarity modeling.
\newblock In {\em In Proceedings of the 2019 Conference on Empirical Methods in
  Natural Language Processing and the 9th International Joint Conference on
  Natural Language Processing {(EMNLP-IJCNLP)}\/} (Hong Kong, China, 2019),
  Association for Computational Linguistics, p.~5369–5380.

\bibitem{Sh18}
{\sc Shen, D., Wang, G., Wang, W., Min, M.~R., Su, Q., Zhang, Y., Li, C.,
  Henao, R., and Carin, L.}
\newblock Baseline needs more love: On simple word-embedding-based models and
  associated pooling mechanisms.
\newblock In {\em Proceedings of the 56th Annual Meeting of the Association for
  Computational Linguistics {(Volume 1: Long Papers)}\/} (Melbourne, Australia,
  2018), Association for Computational Linguistics, pp.~440--450.

\bibitem{Sr14}
{\sc Srivastava, N., Hinton, G., Krizhevsky, A., Sutskever, I., and
  Salakhutdinov, R.}
\newblock Dropout: A simple way to prevent neural networks from overfitting.
\newblock In {\em J. Mach. Learn. Res\/} (2014), {JMLR}.org, p.~1929–1958.

\bibitem{To17}
{\sc Tomar, G.~S., Duque, T., Täckström, O., Uszkoreit, J., and Das, D.}
\newblock Neural paraphrase identification of questions with noisy pretraining.
\newblock In {\em Proceedings of the 1st Workshop on Subword and Character
  Level Models in {NLP}\/} (Copenhagen, Denmark, 2017), Association for
  Computational Linguistics, p.~142–147.

\bibitem{WHF17}
{\sc Wang, Z., Hamza, W., and Florian, R.}
\newblock Bilateral multi-perspective matching for natural language sentences.
\newblock In {\em Proceedings of the Twenty-Sixth International Joint
  Conference on Artificial Intelligence {(IJCAI)}\/} (2017), pp.~4144--4150.

\bibitem{Yang19}
{\sc Yang, R., Zhang, J., Gao, X., Ji, F., and Chen, H.}
\newblock Simple and effective text matching with richer alignment features.
\newblock In {\em Proceedings of the 57th Annual Meeting of the Association for
  Computational Linguistics\/} (Florence, Italy, 2019), Association for
  Computational Linguistics, pp.~4699--4709.

\bibitem{ZZL15}
{\sc Zhang, X., Zhao, J., and LeCun, Y.}
\newblock Character-level convolutional networks for text classification.
\newblock In {\em Proceedings of the 28th International Conference on Neural
  Information Processing Systems - Volume 1\/} (Cambridge, MA, USA, 2015), MIT
  Press, pp.~649--657.

\bibitem{ZW15}
{\sc Zhang, Y., and Wallace, B.}
\newblock A sensitivity analysis of (and practitioners’ guide to)
  convolutional neural networks for sentence classification.
\newblock In {\em arXiv preprint arXiv:1510.03820)\/} (2015).

\end{thebibliography}


\end{document}